\pdfoutput=1
\documentclass{article}

\usepackage[main, final]{neurips_2026}

\usepackage[utf8]{inputenc} 
\usepackage[T1]{fontenc}    
\usepackage{url}            
\usepackage{booktabs}       
\usepackage{amsfonts}       
\usepackage{nicefrac}       
\usepackage{microtype}      
\usepackage{xcolor}         
\usepackage{amsmath, amssymb}
\usepackage{graphicx}
\usepackage{subcaption}
\usepackage{algorithm}
\usepackage{algpseudocode}
\usepackage{multirow}
\usepackage{array}
\usepackage{bm}
\usepackage{hyperref}
\usepackage{cleveref}
\crefname{section}{Sec.}{Secs.}
\Crefname{section}{Sec.}{Secs.}
\crefname{equation}{Eq.}{Eqs.}
\Crefname{equation}{Eq.}{Eqs.}
\crefname{figure}{Fig.}{Figs.}
\Crefname{figure}{Fig.}{Figs.}
\crefname{table}{Tab.}{Tabs.}
\Crefname{table}{Tab.}{Tabs.}
\crefname{appendix}{App.}{Apps.}
\Crefname{appendix}{App.}{Apps.}

\newcommand{\ms}[3][2.3em]{\makebox[#1][r]{#2}\makebox[2.4em][l]{\,\scriptsize$\pm$#3}}

\title{SO(3)-Equivariant Learning on CAD Boundary Representations}

\author{%
  Matteo Ballegeer\thanks{Corresponding author.} \\
  Ghent University \\
  CVAMO, FlandersMake@UGent \\
  \texttt{matteo.ballegeer@ugent.be} \\
  \And
  Dries F. Benoit \\
  Ghent University \\
  CVAMO, FlandersMake@UGent \\
  \texttt{dries.benoit@ugent.be} \\
}
\begin{document}
\maketitle

\begin{abstract}
Learning on boundary representations (B-reps) requires robustness to the pose of a part, since CAD models are not stored in a canonical orientation.
Existing B-rep networks either operate on world-frame coordinates and normals, and thus degrade under rotation, or encode each face in a local reference frame. 
Local frames yield invariance, but reduce inter-face messages to scalars and discard the relative orientation of adjacent faces.
We propose SO3brep, an $\mathrm{SO}(3)$-equivariant graph neural network on the B-rep face-adjacency graph.
Node features are computed in local frames and mapped into a shared frame by Wigner $D$-matrices.
Clebsch--Gordan tensor products couple these features with the directions between adjacent faces, so that relative orientation is preserved throughout message passing and the output remains $\mathrm{SE}(3)$-invariant.
Across three face-segmentation and two classification benchmarks with randomly rotated test parts, SO3brep achieves the highest accuracy among the compared B-rep networks and equivariant point cloud networks, with the fewest parameters.
Its advantage is largest with few labeled parts, a setting typical of industrial CAD.
Code is available at \url{https://github.com/UGent-CVAMO/so3brep}.
\end{abstract}

\section{Introduction}
\label{sec:intro}
Boundary representations (B-reps), the native format of CAD, describe a solid by the parametric faces that bound it~\citep{weiler1986topological}.
Rather than converting the part to a mesh or point cloud, B-rep learning uses the face-adjacency graph~\citep{ansaldi1985geometric} to capture topology, while features on the nodes and edges encode the geometry of the surfaces and their shared curves~\citep{jayaraman2021uvnet}.
With the increasing availability of CAD datasets~\citep{koch2019abc,lambourne2021brepnet}, graph-based B-rep learning has been used for shape classification, face segmentation, manufacturing effort analysis, and B-rep generation~\citep{jayaraman2021uvnet,cao2020graph,lambourne2021brepnet,colligan2022hierarchical,ballegeer2026cad,wu2024aagnet,xu2024brepgen,xu2025autobrep}.

Most B-rep networks, however, read coordinates and surface normals in the world frame~\citep{jayaraman2021uvnet,wu2024aagnet}, so their inputs change when the part is rotated.
Yet the pose of a CAD part is a convention, not a property of the part.
CAD systems leave the choice of up axis to the user, with both SOLIDWORKS and Fusion 360 allowing Y or Z~\citep{solidworks2020upaxis,autodesk2026modeling}, and a part saved inside an assembly carries the assembly's coordinates~\citep{willis2022joinable}.
Benchmarks conceal this sensitivity because their parts share a canonical pose. 
FoV-Net~\citep{ballegeer2026fov} showed that world-frame networks collapse once the test parts are rotated, and achieves rotation invariance by expressing every face in its own local reference frame (LRF).
This invariance has a price: each face is described in its own frame, so the scalar messages it sends discard the orientation relative to the receiving face~\citep{lippmann2025beyond}. 

An equivariant network instead rotates its features with the part and, for an invariant prediction, discards their directional components only at the readout.
When every feature transforms as an irreducible representation (irrep) of $\mathrm{SO}(3)$, messages can carry vectors and tensors in one shared frame, so a face can compare the directions in which its neighbors lie~\citep{thomas2018tensor,weiler20183d}.
In graph networks, such steerable messages have outperformed invariant ones~\citep{brandstetter2022geometric,batzner2022}.
Because the symmetry is built into the layers, the network does not have to learn rotations from augmented copies~\citep{cohen2016group,thomas2018tensor}, and equivariant networks have been shown to be more data-efficient than both non-equivariant~\citep{brehmer2025does} and invariant ones~\citep{batzner2022}.
Data efficiency matters for CAD, where labeled training sets are small, since labels require expert time and many part libraries are proprietary due to intellectual property concerns~\citep{jones2023self,stjepandic2015concurrent}.

We introduce \textbf{SO3brep}, an $\mathrm{SO}(3)$-equivariant graph neural network on the B-rep face-adjacency graph.
Features computed in the local frame of each face's parameterization are rotated into a shared frame and coupled with inter-face directions through Clebsch--Gordan tensor products, making SO3brep, to our knowledge, the first B-rep network with equivariant rather than invariant message passing.
On five CAD benchmarks with randomly rotated test parts, it achieves the highest accuracy with the fewest parameters, with the largest advantage when training data is scarce.

\section{Related work}
\label{sec:related}

\paragraph{Learning on B-reps.}
Learning methods on B-reps take as input the part's face-adjacency graph, in which nodes are faces and edges are the curves they share~\citep{cao2020graph}.
UV-Net~\citep{jayaraman2021uvnet} encodes each face by a CNN on points and normals sampled over its UV domain and each edge by a 1D CNN on its curve; BRepNet~\citep{lambourne2021brepnet} passes messages along oriented coedges.
Later networks add face and edge attributes~\citep{colligan2022hierarchical,wu2024aagnet,zhang2024brepmfr}, graph attention~\citep{lee2023brepgat} or transformers~\citep{dai2025brepformer,zou2025bringing}, or pretrain on unlabeled parts~\citep{jones2023self,lou2023brep}.
The same UV encodings are used for B-rep generation~\citep{xu2024brepgen,xu2025autobrep}.
Some of the added attributes, such as surface type and area, are rotation-invariant, but the UV grids that carry most of the geometry are sampled in the world frame, so the predictions change when the part is rotated~\citep{ballegeer2026fov}.
\citet{heidari2025geometric} survey the field.

\paragraph{Rotation-invariant learning.}
Invariant networks remove the pose from their input through distances and angles~\citep{deng2018ppf,li2021rotation}, a canonical pose from PCA~\citep{li2021closer}, averaging over a set of frames~\citep{puny2022frame}, or local reference frames~\citep{chen2024local}.
\citet{fei2024rotation} survey these approaches for point clouds.
For B-reps, FoV-Net~\citep{ballegeer2026fov} takes the frame of each face from its parameterization and describes the surrounding solid by rays cast in that frame.
Concurrently with this work, EquiCAD~\citep{su2026equicad} encodes face and edge samples with $\mathrm{O}(3)$-equivariant layers and pools them to invariant scalars before message passing.
In both, faces exchange invariant scalars.
Such scalars limit what local message passing can tell apart: descriptors built from distances and angles alone can coincide for distinct structures~\citep{pozdnyakov2022incompleteness}, and invariant message passing fails to distinguish some geometric graphs that equivariant message passing with higher-degree features separates~\citep{joshi2023expressive}.

\paragraph{Equivariant geometric learning.}
Equivariant networks transform their features together with the input in every layer, so directional information is kept throughout.
Group-equivariant and steerable CNNs build the symmetry into convolution~\citep{cohen2016group,cohen2017steerable,cohen2018spherical,weiler20183d,weiler2021coordinate}; tensor field networks and the $\mathrm{SE}(3)$-Transformer~\citep{thomas2018tensor,fuchs2020se} carry it to message passing on irreps coupled by Clebsch--Gordan tensor products~\citep{geiger2022e3nn}.
This design underlies interatomic potentials such as NequIP, MACE, and Equiformer~\citep{batzner2022,batatia2022mace,liao2023equiformer} and equivariant point cloud networks~\citep{fuchs2020se,kim2024continuous}.
Cheaper variants replace the tensor product by scalar-weighted vectors~\citep{satorras2021egnn,deng2021vector} or reduce its cost~\citep{passaro2023reducing,liao2024equiformerv2}.
To our knowledge, no B-rep network passes equivariant messages between faces.
\section{SO3brep}
\label{sec:method}
SO3brep is an $\mathrm{SO}(3)$-equivariant graph neural network on the B-rep face-adjacency graph.
Each face is first described in its own local reference frame (\Cref{sec:lrf}).
Node and edge features built from these frames (\Cref{sec:node_features,sec:edge_features}) are then expressed in one shared world frame, where equivariant message passing combines them (\Cref{sec:gnn}).
An invariant readout makes the prediction independent of the pose.
Positions enter only as differences, making the network translation-invariant as well.

\begin{figure}[t]
\centering
\includegraphics[width=\linewidth]{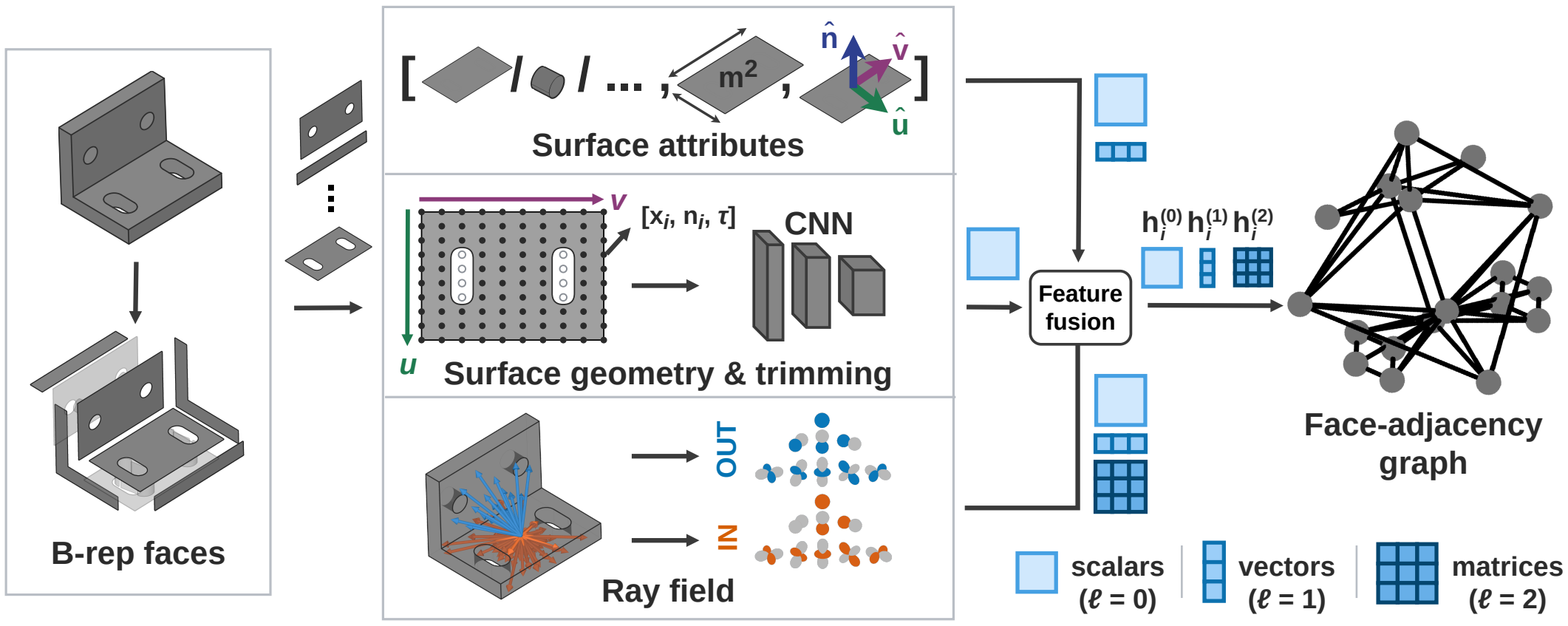}
\caption{\textbf{Face feature construction.}
Each B-rep face is initialized with features from three encoding paths before message passing.
Surface attributes ($\ell{=}0,1$): surface-type and log-area scalars, plus the three LRF axes as $\ell{=}1$ channels.
Surface geometry and trimming ($\ell{=}0$): a CNN on the UV grid, expressed in the LRF.
Ray field ($\ell{=}0,1,2$): a spherical-harmonic (SH) encoding of rays cast over two hemispheres, with ray directions fixed in the LRF.
Feature fusion combines the scalars and projects all paths equivariantly to the GNN input features $\mathbf{h}_i^{(\ell)}$, with $C_\ell$ channels for each degree $\ell \in \{0,1,2\}$.}
\label{fig:node_feat}
\end{figure}

\paragraph{Irreps and spherical harmonics.}
Every feature is an irrep of $\mathrm{SO}(3)$ of degree $\ell \leq \ell_{\max} = 2$, a scalar, vector, or traceless symmetric matrix with $2\ell+1$ components, on which a rotation $\mathbf{R}$ acts by the Wigner $D$-matrix $\mathbf{D}^{(\ell)}(\mathbf{R})$.
A feature with $C$ channels is an array $\mathbf{h}^{(\ell)} \in \mathbb{R}^{(2\ell+1)\times C}$ whose columns are the channels, rotated as $\mathbf{D}^{(\ell)}(\mathbf{R})\,\mathbf{h}^{(\ell)}$; linear maps across channels act from the right, $\mathbf{h}^{(\ell)}\mathbf{W}$, so they commute with this rotation and preserve equivariance.
Directions are encoded by real spherical harmonics (SH) $Y^{(\ell)}(\hat{\mathbf{d}}) \in \mathbb{R}^{2\ell+1}$, which rotate as irreps: $Y^{(\ell)}(\mathbf{R}\hat{\mathbf{d}}) = \mathbf{D}^{(\ell)}(\mathbf{R})\,Y^{(\ell)}(\hat{\mathbf{d}})$.

\subsection{Local reference frames}
\label{sec:lrf}
SO3brep builds its node features in a local reference frame (LRF) per face, a coordinate system anchored at the face that does not depend on the pose of the part.
Each B-rep face is the image of a trimmed domain $\Omega_i \subset \mathbb{R}^2$ under a parametric surface $\mathbf{x}_i$ with parameters $(u,v)$.
We follow~\citet{ballegeer2026fov} and anchor a local frame at the center $(\bar{u}_i,\bar{v}_i)$ of the bounding box of $\Omega_i$, mapped to 3D as $\mathbf{c}_i = \mathbf{x}_i(\bar{u}_i, \bar{v}_i)$.
The CAD kernel provides canonical directions at this point: tangents aligned with $(u,v)$ and a unit normal.
These give an orthonormal basis:
\begin{equation}
\label{eq:lrf}
    \hat{\mathbf{u}}_i =
    \frac{\partial_u \mathbf{x}_i}{\|\partial_u \mathbf{x}_i\|}\bigg|_{(\bar{u}_i,\bar{v}_i)}, \quad
    \hat{\mathbf{n}}_i =
    \frac{\partial_u \mathbf{x}_i \times \partial_v \mathbf{x}_i}
    {\|\partial_u \mathbf{x}_i \times \partial_v \mathbf{x}_i\|}\bigg|_{(\bar{u}_i,\bar{v}_i)}, \quad
    \hat{\mathbf{v}}_i = \hat{\mathbf{n}}_i \times \hat{\mathbf{u}}_i.
\end{equation}
These axes define a rotation matrix $\mathbf{R}_i = [\hat{\mathbf{u}}_i \mid \hat{\mathbf{v}}_i \mid \hat{\mathbf{n}}_i] \in \mathrm{SO}(3)$ whose columns are the frame axes in world coordinates.
Rotating the part by $\mathbf{R}$ maps $\mathbf{R}_i \mapsto \mathbf{R}\mathbf{R}_i$, so the LRF coordinates of any world-frame vector $\mathbf{y}$ do not depend on the pose: $(\mathbf{R}\mathbf{R}_i)^\top \mathbf{R}\mathbf{y} = \mathbf{R}_i^\top \mathbf{y}$.
Thus, $\mathbf{D}^{(\ell)}(\mathbf{R}_i)$ moves features from the LRF into the shared world frame (\Cref{sec:feature_fusion}).

\subsection{Node features}
\label{sec:node_features}
Each face's initial node state $\mathbf{h}_i$ combines three encoding paths (\Cref{fig:node_feat}), computed from its LRF.

\paragraph{Surface attributes.}
We extract analytic scalar attributes directly from the CAD kernel: five surface-type flags (plane, cylinder, cone, sphere, torus), a flag for rational free-form surfaces, and the log-scaled face area, giving $\mathbf{s}_i^{\mathrm{CAD}} \in \mathbb{R}^7$.
The three LRF axes $\hat{\mathbf{u}}_i, \hat{\mathbf{v}}_i, \hat{\mathbf{n}}_i$, expressed in the world frame, also enter the node state as $\ell{=}1$ irreps, so the GNN sees how each face is oriented.

\paragraph{Surface geometry and trimming structure.}
Following UV-Net~\citep{jayaraman2021uvnet}, we sample each face on a regular $10{\times}10$ UV grid, recording at each grid point $(u_p,v_p)$ the 3D position $\mathbf{x}_i(u_p,v_p)$, the surface normal $\mathbf{n}_{ip}$, and a binary trim indicator $\tau_{ip}$, giving a fixed-size representation of the surface geometry regardless of surface type.
All quantities are expressed in the LRF by centering at $\mathbf{c}_i$ and rotating by $\mathbf{R}_i^\top$~\citep{ballegeer2026fov}:
\begin{equation}
    \mathbf{x}'_{ip} = \mathbf{R}_i^\top \bigl(\mathbf{x}_i(u_p,v_p) - \mathbf{c}_i\bigr), \qquad
    \mathbf{n}'_{ip} = \mathbf{R}_i^\top \mathbf{n}_{ip},
\end{equation}
so identical surfaces in different global orientations yield identical representations.
The resulting $10{\times}10{\times}7$ array is processed by a small CNN and pooled to an embedding $\mathbf{s}_i^{\mathrm{UV}}$ of $\ell{=}0$ features.

\paragraph{Ray field.}
To describe the geometry around each face, we follow~\citet{ballegeer2026fov} and cast $M{=}72$ rays over each of two Fibonacci hemisphere grids defined in the LRF.
The outward hemisphere, centered on $\hat{\mathbf{n}}_i$, probes the surrounding geometry; the inward one, centered on $-\hat{\mathbf{n}}_i$, probes the solid interior.
Each ray returns whether it hits the part, the hit distance, and the cosine between the ray and the surface normal at the hit point.
Inside a closed solid every inward ray hits, so for inward rays the hit indicator is replaced by the proximity $1/(1+\text{distance})$, which is large behind thin walls.
A ray that misses returns zeros.
Each face thus carries six signals, three per hemisphere; we write $g_{isk}$ for the value of signal $s$ on ray $k$, $\hat{\mathbf{d}}_{sk}$ for the direction of that ray in the LRF (shared by the three signals of a hemisphere), and $\bar{g}_{is}$ for the mean of the signal over its hemisphere.
Because the directions are fixed, their SH encoding is a fixed linear map:
\begin{equation}
    \mathbf{f}_{is}^{(0)} = \bar{g}_{is},
    \qquad
    \mathbf{f}_{is}^{(\ell)}
    =
    \frac{1}{M}
    \sum_{k=1}^{M}
    \bigl(g_{isk} - \bar{g}_{is}\bigr)\, Y^{(\ell)}(\hat{\mathbf{d}}_{sk}),
    \quad \ell \in \{1,2\}.
\end{equation}
On a hemisphere $Y^{(\ell)}$ does not average to zero, so centering is needed for the $\ell{>}0$ coefficients to describe only how the signal varies with direction.
This gives six degree-$\ell$ irreps $\mathbf{f}_i^{(\ell)} \in \mathbb{R}^{(2\ell+1) \times 6}$: $\ell{=}0$ holds the mean of each signal, $\ell{=}1$ its dominant direction, and $\ell{=}2$ its quadrupolar contrast.
The $\ell{=}0$ coefficients and the norms of the $\ell{>}0$ coefficients feed a scalar MLP.

\paragraph{Feature fusion.}%
\label{sec:feature_fusion}
The three scalar sources (analytic CAD attributes $\mathbf{s}_i^{\mathrm{CAD}}$, UV-grid embeddings $\mathbf{s}_i^{\mathrm{UV}}$, and the ray scalars from the MLP above) are each projected to width $C_0$ and fused by an MLP, producing the scalar node representation $\mathbf{h}_i^{(0)} \in \mathbb{R}^{C_0}$.
The ray coefficients $\mathbf{f}_i^{(\ell)}$ with $\ell{>}0$ are expressed in each face's own LRF, so the same coefficient describes a different direction on every face.
We rotate them into the world frame with the Wigner $D$-matrix of the LRF~\citep{geiger2022e3nn}:%
\abovedisplayskip=7pt plus 2pt\belowdisplayskip=7pt plus 2pt 
\begin{equation}
    \tilde{\mathbf{f}}_i^{(\ell)} = \mathbf{D}^{(\ell)}(\mathbf{R}_i)\,\mathbf{f}_i^{(\ell)}.
\end{equation}
The coefficients are fixed relative to the face and the LRF turns with the part, so $\tilde{\mathbf{f}}_i^{(\ell)}$ turns with the part as well.
A linear layer per degree maps them, together with the LRF axes, to the vector and matrix node features $\mathbf{h}_i^{(1)}$ and $\mathbf{h}_i^{(2)}$.

\subsection{Edge features}
\label{sec:edge_features}
Each edge of the face-adjacency graph carries two descriptors of the faces it joins: directional features and rotation-invariant scalars.
The directional features enter the Clebsch--Gordan (CG) tensor product, and the scalars set the strength of each message through the radial network (\Cref{sec:gnn}).

\begin{figure}[t]
\centering
\includegraphics{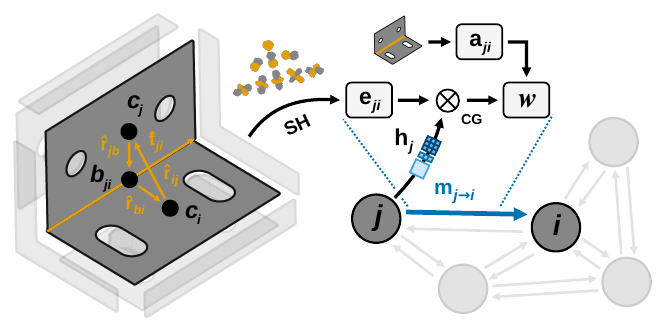}
\caption{\textbf{Equivariant message passing.}
Four unit vectors join the face centers $\mathbf{c}_i$, $\mathbf{c}_j$ and the shared edge: $\hat{\mathbf{r}}_{ij}$, $\hat{\mathbf{r}}_{jb}$, $\hat{\mathbf{r}}_{bi}$, and the edge chord $\hat{\mathbf{t}}_{ji}$.
Their SH $\mathbf{e}_{ji}$ are coupled with the sender's irreps $\mathbf{h}_j$ by CG tensor products, so each message carries the relative directions of the two faces.
A radial network maps the invariant edge scalars $\mathbf{a}_{ji}$ and the center distance to the gates of each coupling path.
}
\label{fig:equivariant_mp}
\end{figure}

\paragraph{Directional edge features.}
Adjacent faces $i$ and $j$ are joined by an edge in each direction; along the edge $ji$, the sender $j$ passes a message to the receiver $i$.
We summarize the relative orientation of the two faces with four world-frame unit vectors (\Cref{fig:equivariant_mp}), built from the face centers $\mathbf{c}_i$, $\mathbf{c}_j$, the parametric midpoint $\mathbf{b}_{ji}$ of their shared boundary curve, and its endpoints $\mathbf{p}_{ji}$, $\mathbf{p}'_{ji}$:
\begin{equation}
\label{eq:edge_dirs}
    \hat{\mathbf{r}}_{ij} = \frac{\mathbf{c}_j - \mathbf{c}_i}{d_{ji}}, \quad
    \hat{\mathbf{r}}_{jb} = \frac{\mathbf{b}_{ji} - \mathbf{c}_j}{\|\mathbf{b}_{ji} - \mathbf{c}_j\|}, \quad
    \hat{\mathbf{r}}_{bi} = \frac{\mathbf{c}_i - \mathbf{b}_{ji}}{\|\mathbf{c}_i - \mathbf{b}_{ji}\|}, \quad
    \hat{\mathbf{t}}_{ji} = \frac{\mathbf{p}'_{ji} - \mathbf{p}_{ji}}{\|\mathbf{p}'_{ji} - \mathbf{p}_{ji}\|},
\end{equation}
with $d_{ji} = \|\mathbf{c}_j - \mathbf{c}_i\|$. The subscripts of $\hat{\mathbf{r}}$ give its direction: $\hat{\mathbf{r}}_{ij}$ points from the receiver center $\mathbf{c}_i$ to the sender center $\mathbf{c}_j$; $\hat{\mathbf{r}}_{jb}$ and $\hat{\mathbf{r}}_{bi}$ follow the message from $\mathbf{c}_j$ through the shared boundary to $\mathbf{c}_i$; and $\hat{\mathbf{t}}_{ji}$ is the chord of the boundary curve, signed by the curve's orientation.
Where the two points of a difference coincide, as at the endpoints of a closed curve, the vector is set to zero, with $Y^{(\ell)}(\mathbf{0}) = \mathbf{0}$ for $\ell{>}0$.
Their SH form the four channels of the edge irreps,
\begin{equation}
    \mathbf{e}_{ji}^{(\ell)} = \bigl[\,Y^{(\ell)}(\hat{\mathbf{r}}_{ij}),\; Y^{(\ell)}(\hat{\mathbf{r}}_{jb}),\; Y^{(\ell)}(\hat{\mathbf{r}}_{bi}),\; Y^{(\ell)}(\hat{\mathbf{t}}_{ji})\,\bigr] \in \mathbb{R}^{(2\ell+1) \times 4}, \quad \ell \in \{1,2\},
\end{equation}
and the constant $\mathbf{e}_{ji}^{(0)} = Y^{(0)}$ is kept once.

\paragraph{Invariant edge scalars.}
The scalar descriptor $\mathbf{a}_{ji} \in \mathbb{R}^9$ holds the cosine $\hat{\mathbf{n}}_j \cdot \hat{\mathbf{n}}_i$ between the face normals and eight attributes of the shared curve: two convexity flags (concave, convex; both zero for smooth edges), the arc length (log-compressed), whether the curve is closed, and four curve-type flags (circle, non-rational B-spline, rational B-spline, line).
The center distance is expanded in 16 Gaussian radial basis functions, $\mathrm{rbf}(d_{ji}) \in \mathbb{R}^{16}$~\citep{schutt2017schnet}.

\subsection{Equivariant message passing}
\label{sec:gnn}
Node features are updated by $L$ layers, each made of two residual blocks: equivariant message passing and an equivariant feed-forward network.
Both blocks start with an RMS normalization~\citep{zhang2019root}, taken per irrep block so that it is equivariant~\citep{liao2023equiformer}.

\paragraph{Message construction.}
The CG tensor product $\otimes_{\mathrm{CG}}$~\citep{thomas2018tensor,geiger2022e3nn} couples a degree-$\ell_1$ sender feature $\mathbf{h}_j^{(\ell_1)}$ with a degree-$\ell_2$ edge irrep $\mathbf{e}_{ji}^{(\ell_2)}$ into a degree-$\ell_3$ output, which is possible for every pair in $\mathcal{I}_{\ell_3} = \{(\ell_1,\ell_2) \in \{0,\ldots,\ell_{\max}\}^2 : |\ell_1-\ell_2| \leq \ell_3 \leq \ell_1+\ell_2\}$.
Every channel of $\mathbf{h}_j^{(\ell_1)}$ is coupled with every channel of $\mathbf{e}_{ji}^{(\ell_2)}$.
For instance, a sender vector and an edge direction ($\ell_1{=}\ell_2{=}1$) couple to their dot product ($\ell_3{=}0$), their cross product ($\ell_3{=}1$), and a traceless symmetric matrix ($\ell_3{=}2$).
The message of degree $\ell_3$ gates these products and mixes them:
\begin{equation}
\label{eq:tp}
\mathbf{m}_{j \to i}^{(\ell_3)} =
\left(
\bigoplus_{(\ell_1,\ell_2)\in \mathcal{I}_{\ell_3}}
\left[
\left(
\mathbf{h}_j^{(\ell_1)} \otimes_{\mathrm{CG}} \mathbf{e}_{ji}^{(\ell_2)}
\right)^{(\ell_3)}
\odot\, w^{(\ell_1,\ell_2,\ell_3)}(\mathbf{a}_{ji}, \mathrm{rbf}(d_{ji}))
\right]
\right)
\mathbf{W}^{(\ell_3)},
\end{equation}
where $w$ is the radial network, an MLP on the invariant edge scalars, $\odot$ multiplies each channel by its gate, $\bigoplus$ concatenates the coupling paths along the channel axis, and $\mathbf{W}^{(\ell_3)}$ mixes channels.
All CG coefficients are fixed; the learnable parameters are $\mathbf{W}^{(\ell_3)}$ and the radial network.

\paragraph{Aggregation with scalar attention.}
Messages from the neighborhood $\mathcal{N}(i)$ are aggregated using learned attention weights.
For each of $A$ heads, a scalar query $q_i^{(a)}$ and key $k_j^{(a)}$ are linear functions of the $\ell{=}0$ channels of nodes $i$ and $j$, giving attention weights $\alpha_{ji}^{(a)} = \mathrm{softmax}_{j \in \mathcal{N}(i)}\bigl(q_i^{(a)} k_j^{(a)} / \sqrt{C_0}\bigr)$.
Each head weights a disjoint, equal-width slice of the message channels within each irrep block.
The aggregated features are then linearly mixed and combined with a residual connection:
\begin{equation}
    \mathbf{h}_i^{(\ell)\prime} =
    \mathbf{h}_i^{(\ell)} +
    \left(
    \sum_{j \in \mathcal{N}(i)}
    \bigoplus_{a=1}^{A}
    \alpha_{ji}^{(a)}\,
    \mathbf{m}_{j \to i}^{(a,\ell)}
    \right)
    \mathbf{V}^{(\ell)},
\end{equation}
where $\mathbf{m}_{j\to i}^{(a,\ell)}$ denotes the $a$-th head slice of $\mathbf{m}_{j\to i}^{(\ell)}$, and $\mathbf{V}^{(\ell)}$ mixes channels.

\paragraph{Equivariant feed-forward block.}
A second residual block transforms each node equivariantly:%
\abovedisplayskip=7pt plus 2pt\belowdisplayskip=7pt plus 2pt 
\begin{equation}
\mathbf{h}_i'' = \mathbf{h}_i' + \mathrm{FFN}(\mathbf{h}_i').
\end{equation}
The FFN applies a linear map across channels, the gated nonlinearity of~\citet{weiler20183d}, and a second linear map.
The nonlinearity passes $\ell{=}0$ channels through SiLU and multiplies each $\ell{>}0$ channel by the sigmoid of an extra $\ell{=}0$ output of the first map~\citep{geiger2022e3nn}.
Every layer is equivariant: edge SH rotate with the part, CG products are equivariant, all gates and attention weights are invariant scalars on whole channels, and linear maps act across channels only.

\paragraph{Invariant readout.}
The $\ell{=}0$ components of the final node states are projected via a two-layer MLP to per-node descriptors $\mathbf{z}_i \in \mathbb{R}^{d_{\mathrm{out}}}$; higher-degree components are discarded to ensure an invariant output.
The graph descriptor $\mathbf{z}_G$ concatenates the mean and max pooling of all $\mathbf{z}_i$.
For classification, $\mathbf{z}_G$ is passed to an MLP head.
For segmentation, each $\mathbf{z}_i$ is concatenated with $\mathbf{z}_G$ before a classification head that outputs per-face class probabilities.

\section{Experiments}
\label{sec:experiments}
We evaluate SO3brep on three face-segmentation and two classification benchmarks in \Cref{sec:exp_seg,sec:exp_cls}, compare message functions on identical inputs (\Cref{sec:exp_matched}), measure data efficiency (\Cref{sec:data_efficiency}), check equivariance numerically (\Cref{sec:exp_equiv}), and close with ablations (\Cref{sec:ablation}).

\paragraph{Datasets.}
For face segmentation, \textbf{MFCAD++}~\citep{colligan2022hierarchical} (59.6K parts, 25 classes) and \textbf{MFTRCAD}~\citep{xia2024machining} (28.6K parts, 27 classes) label each face with a machining feature; MFTRCAD adds rotational features and multi-layer Boolean operations.
\textbf{Fusion 360}~\citep{lambourne2021brepnet} (35.7K parts, 8 classes) labels each face with the modeling operation that created it.
For classification, \textbf{SolidLetters}~\citep{jayaraman2021uvnet} contains 96K solids of 26 letters in 2,002 fonts, and \textbf{BenDFM}~\citep{ballegeer2026bendfm} contains 14K sheet-metal parts labeled by whether they can be bent without the part colliding with the tooling.

\paragraph{Baselines.}
We compare against four methods: (i)~\textbf{UV-Net}~\citep{jayaraman2021uvnet}, which encodes coordinates and normals in the world frame; (ii)~\textbf{FoV-Net}~\citep{ballegeer2026fov}, an $\mathrm{SE}(3)$-invariant B-rep network built on per-face frames; (iii)~\textbf{VN-PointNet}~\citep{deng2021vector} and (iv)~\textbf{CSEConv}~\citep{kim2024continuous}, two equivariant point cloud networks trained on 1,024 points sampled by farthest point sampling on the faces.
CSEConv supports classification only.

\paragraph{Evaluation protocol and metrics.}
All models are trained on the original orientations of each dataset without rotation augmentation and tested twice: on the original test set (Original) and with every test part rotated about the three coordinate axes by random angles between 0 and $2\pi$ ($\mathrm{SO}(3)$).
Segmentation is scored by per-face accuracy and mean intersection-over-union (mIoU) over classes, classification by accuracy. 
Point cloud baselines are scored per face by a majority vote over the points sampled on that face.
We report mean\,$\pm$\,std over five seeds.

\paragraph{Implementation.}
Node features have $C_0{=}64$, $C_1{=}16$, and $C_2{=}8$ channels of degree $\ell{=}0,1,2$, with $L{=}3$ equivariant GNN layers and $A{=}8$ attention heads; the readout uses $d_{\mathrm{out}}{=}128$.
All models are trained with cross-entropy loss, batch size 64, and early stopping on validation loss (patience 25) on a single NVIDIA A5000 GPU.
SO3brep uses AdamW (learning rate $10^{-3}$, weight decay 0.05) with a 5-epoch linear warm-up and cosine decay to $2{\times}10^{-5}$, feature dropout $p{=}0.15$, and equivariant channel dropout $p{=}0.10$.
UV-Net uses Adam, and FoV-Net and the point cloud baselines use AdamW (weight decay 0.01), all with learning rate $10^{-3}$.
\Cref{app:cost} compares computational cost and reports graph sizes.

\subsection{Face segmentation}
\label{sec:exp_seg}

\Cref{tab:seg} shows that SO3brep has the highest accuracy and mIoU on all three segmentation benchmarks, with and without test rotations: 99.45\% accuracy on MFCAD++, 92.80\% on MFTRCAD, and 95.11\% on Fusion 360.
Its original and rotated scores coincide because its output is invariant.
UV-Net loses 19--58 accuracy points under rotation, since its input coordinates and normals are tied to the canonical pose.
FoV-Net is invariant as well and is the strongest baseline; SO3brep exceeds it by 0.4, 2.5, and 3.4 accuracy points and by 1.2, 6.7, and 8.4 mIoU points on MFCAD++, MFTRCAD, and Fusion 360.
VN-PointNet, which sees 1024 sampled points and labels each face by a vote, reaches 17.5--31.4\% mIoU against 71.8--98.5\% for FoV-Net and SO3brep; \citet{jayaraman2021uvnet} report the same ordering for non-equivariant networks.

\begin{table}[t]
\centering
\caption{Face segmentation on original and randomly rotated test parts.
All models are trained on the original orientations.
Accuracy and mIoU in \%, mean\,$\pm$\,std over five random seeds.
Best in \textbf{bold}, second best \underline{underlined}.}
\label{tab:seg}
\footnotesize
\begin{tabular*}{\textwidth}{@{\extracolsep{\fill}}l c cc cc}
\toprule
& & \multicolumn{2}{c}{Acc.} & \multicolumn{2}{c}{mIoU} \\
\cmidrule(lr){3-4} \cmidrule(lr){5-6}
Model & Params (M) & Original & $\mathrm{SO}(3)$ & Original & $\mathrm{SO}(3)$ \\
\midrule
\multicolumn{6}{@{}l}{\textit{MFCAD++}} \\
UV-Net & 1.37 & \ms{98.62}{0.05} & \ms{41.09}{6.38} & \ms{96.02}{0.15} & \ms{20.66}{3.04} \\
FoV-Net & 0.59 & \ms{\underline{99.05}}{0.06} & \ms{\underline{99.05}}{0.06} & \ms{\underline{97.27}}{0.20} & \ms{\underline{97.27}}{0.20} \\
VN-PointNet & 1.08 & \ms{83.57}{1.61} & \ms{83.57}{1.61} & \ms{31.38}{3.51} & \ms{31.38}{3.51} \\
\textbf{SO3brep} & \textbf{0.41} & \ms{\textbf{99.45}}{0.02} & \ms{\textbf{99.45}}{0.02} & \ms{\textbf{98.47}}{0.12} & \ms{\textbf{98.47}}{0.12} \\
\midrule
\multicolumn{6}{@{}l}{\textit{MFTRCAD}} \\
UV-Net & 1.37 & \ms{89.54}{0.18} & \ms{38.29}{2.01} & \ms{68.08}{0.57} & \ms{18.24}{0.70} \\
FoV-Net & 0.59 & \ms{\underline{90.28}}{0.26} & \ms{\underline{90.28}}{0.26} & \ms{\underline{71.84}}{0.40} & \ms{\underline{71.84}}{0.40} \\
VN-PointNet & 1.08 & \ms{77.12}{1.09} & \ms{77.12}{1.09} & \ms{17.52}{1.35} & \ms{17.52}{1.35} \\
\textbf{SO3brep} & \textbf{0.41} & \ms{\textbf{92.80}}{0.18} & \ms{\textbf{92.80}}{0.18} & \ms{\textbf{78.54}}{0.52} & \ms{\textbf{78.54}}{0.52} \\
\midrule
\multicolumn{6}{@{}l}{\textit{Fusion 360}} \\
UV-Net & 1.36 & \ms{87.59}{0.16} & \ms{68.60}{1.53} & \ms{65.92}{0.99} & \ms{37.16}{0.98} \\
FoV-Net & 0.59 & \ms{\underline{91.76}}{0.34} & \ms{\underline{91.76}}{0.34} & \ms{\underline{73.69}}{1.44} & \ms{\underline{73.69}}{1.44} \\
VN-PointNet & 1.08 & \ms{74.72}{1.62} & \ms{74.72}{1.62} & \ms{28.49}{2.23} & \ms{28.49}{2.23} \\
\textbf{SO3brep} & \textbf{0.40} & \ms{\textbf{95.11}}{0.12} & \ms{\textbf{95.11}}{0.12} & \ms{\textbf{82.05}}{0.50} & \ms{\textbf{82.05}}{0.50} \\
\bottomrule
\end{tabular*}
\end{table}

\subsection{Classification}
\label{sec:exp_cls}

\begin{table}[t]
\centering
\caption{Shape classification on original and randomly rotated test parts.
All models are trained on the original orientations.
Accuracy in \%, mean\,$\pm$\,std over five random seeds.
Best in \textbf{bold}, second best \underline{underlined}.}
\label{tab:cls}
\footnotesize
\begin{tabular*}{\textwidth}{@{\extracolsep{\fill}}l c cc cc}
\toprule
& & \multicolumn{2}{c}{SolidLetters} & \multicolumn{2}{c}{BenDFM} \\
\cmidrule(lr){3-4} \cmidrule(lr){5-6}
Model & Params (M) & Original & $\mathrm{SO}(3)$ & Original & $\mathrm{SO}(3)$ \\
\midrule
UV-Net & 1.29 & \ms{\textbf{97.36}}{0.14} & \ms{9.22}{0.70} & \ms{75.50}{0.45} & \ms{56.53}{1.12} \\
FoV-Net & 0.59 & \ms{96.38}{0.14} & \ms{\underline{96.38}}{0.14} & \ms{\underline{81.13}}{0.18} & \ms{\underline{81.13}}{0.18} \\
VN-PointNet & 0.81 & \ms{85.81}{0.40} & \ms{85.81}{0.40} & \ms{73.55}{0.18} & \ms{73.55}{0.18} \\
CSEConv & 2.09 & \ms{87.48}{0.18} & \ms{87.48}{0.18} & \ms{76.69}{0.77} & \ms{76.69}{0.77} \\
\midrule
\textbf{SO3brep} & \textbf{0.36} & \ms{\underline{97.03}}{0.13} & \ms{\textbf{97.03}}{0.13} & \ms{\textbf{81.68}}{0.79} & \ms{\textbf{81.68}}{0.79} \\
\bottomrule
\end{tabular*}
\end{table}

Under rotation, SO3brep has the highest accuracy on both datasets (\Cref{tab:cls}): 97.03\% on SolidLetters and 81.68\% on BenDFM.
On BenDFM, its margin over FoV-Net ($+0.55$ points) is within the seed spread.
On SolidLetters, UV-Net is 0.33 points more accurate on aligned parts but drops to 9.22\% under rotation.
Part of the SolidLetters labels depends on pose: the extruded letters $b$, $d$, $p$, and $q$ are close to rotations of one another, so an $\mathrm{SO}(3)$-invariant model can separate them only by font details, while a pose-sensitive model trained and tested on aligned parts can read their orientation.
\citet{fuchs2020se} observe the same on the gravity-aligned ScanObjectNN, where giving their model the ``up'' direction raises its accuracy.

\subsection{Message functions on identical inputs}
\label{sec:exp_matched}
SO3brep introduces both new input features and equivariant message passing.
To separate the two, we keep SO3brep's inputs fixed and vary only the message function (\Cref{tab:matched}).
The \textit{non-equivariant} variants replace the CG tensor product in \Cref{fig:equivariant_mp} with attention or concatenation over the same world-frame features.
The \textit{invariant} variant skips the rotation by the Wigner $D$-matrices: $\ell{>}0$ node features stay in their LRFs, directional edge features are expressed in the sender's LRF, and, with no shared frame left for CG coupling, features are concatenated.
It is the closest LRF-based counterpart of SO3brep and therefore the direct test of equivariant against invariant message passing.

\paragraph{Results.}
With the full training set and aligned parts, all four message functions are within 1.5 mIoU points of one another.
The non-equivariant variants lose $7$--$18$ accuracy and $19$--$21$ mIoU points under rotation, since combining world-frame spherical-harmonic features without CG tensor products breaks equivariance; they also use $1.7$--$2\times$ the parameters.
Against the invariant message function, SO3brep is $0.1$--$0.3$ accuracy points and $1.1$--$1.5$ mIoU points higher at full data with $40\%$ fewer parameters.
With 100 training parts the difference grows to $1.9$ accuracy and $3.7$ mIoU points on Fusion 360 and to $13.5$ and $13.1$ on MFTRCAD.
Under rotation, SO3brep's lead over the non-equivariant variants with 100 training parts is $4.5$--$7.0$ accuracy and $6.6$--$12.7$ mIoU points on Fusion 360 and $12.0$--$15.8$ and $10.5$--$14.1$ on MFTRCAD.

\begin{table}[t]
\centering
\caption{Controlled comparison of message functions on identical input features.
FoV-Net uses its own architecture and features and is shown for reference.
Accuracy and mIoU in \%, mean\,$\pm$\,std over five random seeds (for 100 training parts, over five random training subsets).
Best of the four message functions in \textbf{bold}.}
\label{tab:matched}
\footnotesize
\setlength{\tabcolsep}{3pt}
\begin{tabular*}{\textwidth}{@{\extracolsep{\fill}}l c cc cc cc}
\toprule
& & \multicolumn{4}{c}{Full training set} & \multicolumn{2}{c}{100 training parts} \\
\cmidrule(lr){3-6} \cmidrule(lr){7-8}
& & \multicolumn{2}{c}{Original} & \multicolumn{2}{c}{$\mathrm{SO}(3)$} & \multicolumn{2}{c}{$\mathrm{SO}(3)$} \\
\cmidrule(lr){3-4} \cmidrule(lr){5-6} \cmidrule(lr){7-8}
Message function & Params (M) & Acc. & mIoU & Acc. & mIoU & Acc. & mIoU \\
\midrule
\multicolumn{8}{@{}l}{\textit{Fusion 360}} \\
\textbf{SO3brep} (CG) & \textbf{0.40} & \ms{95.11}{0.12} & \ms{\textbf{82.05}}{0.50} & \ms{\textbf{95.11}}{0.12} & \ms{\textbf{82.05}}{0.50} & \ms{\textbf{83.45}}{1.51} & \ms{\textbf{48.53}}{4.26} \\
Attention (non-eq.) & 0.82 & \ms{\textbf{95.31}}{0.10} & \ms{81.84}{0.51} & \ms{87.95}{0.27} & \ms{61.89}{0.82} & \ms{79.00}{1.52} & \ms{41.92}{2.43} \\
Concat. (non-eq.) & 0.68 & \ms{95.08}{0.25} & \ms{81.09}{1.02} & \ms{87.81}{1.11} & \ms{62.14}{1.86} & \ms{76.44}{1.36} & \ms{35.86}{2.65} \\
Invariant (LRF) & 0.68 & \ms{94.78}{0.14} & \ms{80.56}{1.06} & \ms{94.78}{0.14} & \ms{80.56}{1.06} & \ms{81.51}{1.18} & \ms{44.83}{2.42} \\
\textcolor{gray}{FoV-Net (ref.)} & \textcolor{gray}{0.59} & \textcolor{gray}{\ms{91.76}{0.34}} & \textcolor{gray}{\ms{73.69}{1.44}} & \textcolor{gray}{\ms{91.76}{0.34}} & \textcolor{gray}{\ms{73.69}{1.44}} & \textcolor{gray}{\ms{71.82}{1.32}} & \textcolor{gray}{\ms{33.74}{1.45}} \\
\midrule
\multicolumn{8}{@{}l}{\textit{MFTRCAD}} \\
\textbf{SO3brep} (CG) & \textbf{0.41} & \ms{92.80}{0.18} & \ms{\textbf{78.54}}{0.52} & \ms{\textbf{92.80}}{0.18} & \ms{\textbf{78.54}}{0.52} & \ms{\textbf{81.58}}{0.75} & \ms{\textbf{43.01}}{1.65} \\
Attention (non-eq.) & 0.82 & \ms{\textbf{92.87}}{0.10} & \ms{78.10}{0.45} & \ms{75.10}{2.54} & \ms{58.67}{2.36} & \ms{69.55}{2.32} & \ms{32.56}{2.02} \\
Concat. (non-eq.) & 0.68 & \ms{92.76}{0.24} & \ms{77.81}{0.73} & \ms{75.28}{1.61} & \ms{56.76}{1.27} & \ms{65.75}{3.91} & \ms{28.92}{3.45} \\
Invariant (LRF) & 0.68 & \ms{92.69}{0.06} & \ms{77.42}{0.40} & \ms{92.69}{0.06} & \ms{77.42}{0.40} & \ms{68.11}{0.90} & \ms{29.87}{1.19} \\
\textcolor{gray}{FoV-Net (ref.)} & \textcolor{gray}{0.59} & \textcolor{gray}{\ms{90.28}{0.26}} & \textcolor{gray}{\ms{71.84}{0.40}} & \textcolor{gray}{\ms{90.28}{0.26}} & \textcolor{gray}{\ms{71.84}{0.40}} & \textcolor{gray}{\ms{57.05}{2.38}} & \textcolor{gray}{\ms{18.36}{2.48}} \\
\bottomrule
\end{tabular*}
\end{table}

\subsection{Data efficiency}\label{sec:data_efficiency}
\begin{figure}[t]
\centering
\includegraphics[width=\linewidth]{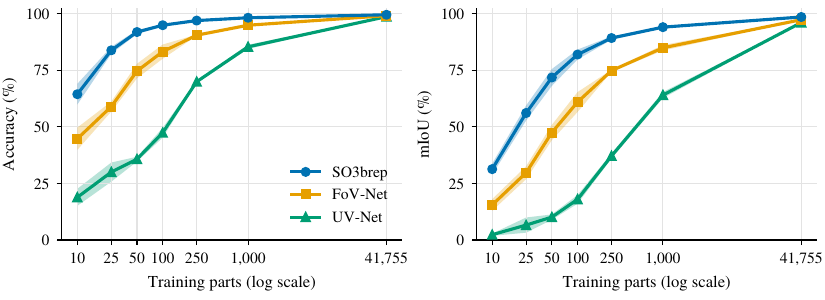}
\caption{\textbf{Data efficiency.} Learning curves for MFCAD++ segmentation versus training size (log scale). 
SO3brep is above UV-Net and FoV-Net at every training-set size, with the largest gap at the smallest sizes.
Shaded areas give the standard deviation over five random splits.}
\label{fig:lc}
\end{figure}

Following~\citet{jones2023self,ballegeer2026fov}, we evaluate data efficiency on the aligned MFCAD++ segmentation dataset by training on smaller data subsets, using five random splits per subset size (\Cref{fig:lc}).
SO3brep is the most accurate model at every training-set size, and the gap is largest with few samples: with 10 training parts it is about 20 points more accurate than FoV-Net and 45 points more accurate than UV-Net, and with 50 parts it reaches 91.8\%.
Part of this gap comes from the directional input features SO3brep enables, as in \Cref{tab:matched}. 
Industrial CAD training sets are often small, since labels require expert time and many part libraries are proprietary~\citep{jones2023self,stjepandic2015concurrent}, so the accuracy reached with few labeled parts decides whether a B-rep network can be used in practice.

\subsection{Equivariance validation}
\label{sec:exp_equiv}

We measure the relative equivariance error~\citep{fuchs2020se} on the vector ($\ell{=}1$) features of the last equivariant layer of each equivariant model,
\begin{equation}
\label{eq:equiv_error}
\Delta_{\mathrm{EQ}} = \frac{\|\Phi(\mathbf{R}\cdot\mathcal{X}) - \rho(\mathbf{R})\,\Phi(\mathcal{X})\|_2}{\|\Phi(\mathbf{R}\cdot\mathcal{X})\|_2},
\qquad \rho(\mathbf{R}) = \mathbf{D}^{(1)}(\mathbf{R}),
\end{equation}
where $\Phi$ is the feature extractor, $\mathcal{X}$ the input, and $\mathbf{R}$ a random rotation; we average over 1,000 SolidLetters inputs and 100 rotations per input.
All three equivariant models are equivariant up to floating-point precision: $\Delta_{\mathrm{EQ}} = (3.46\pm4.91)\times10^{-6}$ for SO3brep, $(9.83\pm9.80)\times10^{-6}$ for VN-PointNet, and $(1.00\pm0.36)\times10^{-7}$ for CSEConv. 


\subsection{Ablation studies}
\label{sec:ablation}
\begin{figure}[t]
\centering
\includegraphics[width=\linewidth]{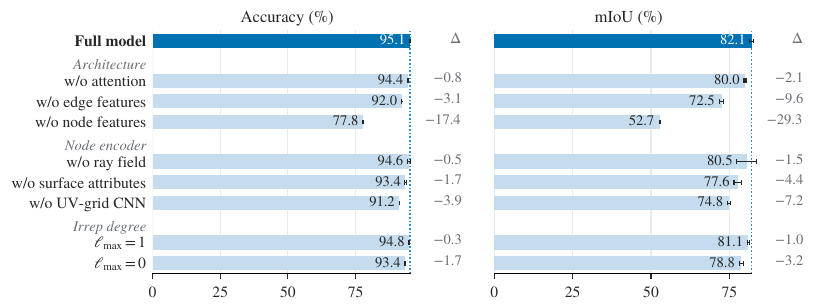}
\caption{\textbf{Ablations on Fusion 360 face segmentation.}
Each row removes or reduces one component of SO3brep; bars give accuracy and mIoU (\%, mean\,$\pm$\,std over five random seeds), and $\Delta$ the change from the full model in percentage points.}
\label{fig:ablation_fusion360}
\end{figure}

\Cref{fig:ablation_fusion360} ablates SO3brep on Fusion 360 segmentation.
The architecture variants replace the node encodings with learned constants (\textit{w/o node features}), replace the edge scalars and directional features with constants so that messages carry only topology (\textit{w/o edge features}), or sum over neighbors instead of attending (\textit{w/o attention}).
The node-encoder variants remove one of the three face feature streams each, and the irrep degree ablations lower $\ell_{\max}$ to 1 and 0.
Node features matter most: without them, accuracy drops by 17.4 points and mIoU by 29.3.
Edge features follow (3.1 and 9.6 points), then attention (0.8 and 2.1).
Among the node encoders, the UV-grid CNN contributes most (3.9 and 7.2) and the ray field least (0.5 and 1.5).
Lowering $\ell_{\max}$ costs 0.3 and 1.0 points at $\ell_{\max}{=}1$ and 1.7 and 3.2 at $\ell_{\max}{=}0$, where only scalar features remain.

\section{Conclusion}
\label{sec:conclusion}

We introduced SO3brep, an $\mathrm{SO}(3)$-equivariant network for CAD B-reps.
Node features are built in per-face frames and rotated into a shared frame with Wigner $D$-matrices, and messages couple them with the directions between faces through CG tensor products, so the network keeps the relative orientations of faces while its output is $\mathrm{SE}(3)$-invariant.
On five benchmarks with randomly rotated test parts, it has the highest accuracy among the compared B-rep and equivariant point cloud methods, with the fewest parameters and the largest margin when training data is scarce.
We hope SO3brep encourages further work on equivariant geometric deep learning for CAD.

\paragraph{Limitations.}
The edge features are limited to four unit vectors and nine scalars; richer descriptors between surfaces are possible.
In-plane tangent directions are defined only up to rotation about $\hat{\mathbf{n}}_i$, making the $u/v$ axes and edge orientations dependent on parameterization conventions and potentially inconsistent across cross-system B-rep exports.
Operations that keep the solid but change its B-rep, such as splitting faces by axis-aligned planes or re-parameterizing surfaces as NURBS, also lower the accuracy of B-rep networks~\citep{jiang2026learn}.
CG tensor products scale as $O(\ell_{\max}^{6})$ in the maximum degree, which limits $\ell_{\max}$ in practice~\citep{passaro2023reducing,liao2024equiformerv2}.
Predictions are per face, so continuous surface signals such as pressure fields are out of scope for the current architecture.
Finally, with full training data the current benchmarks are close to saturated: all message functions in \Cref{tab:matched} are within 1.5 mIoU points on aligned parts, so these benchmarks show little of the expressivity advantage of equivariance over invariance.

\paragraph{Future work.}
Equivariant architectures enable tasks whose outputs must rotate with the input, such as predicting a machining approach direction, a drilling axis, or a part-to-part mating transform~\citep{willis2022joinable,wu2023leveraging}; invariant architectures cannot express these, and public CAD benchmarks for them are still missing.
On the architecture side, we aim to reduce CAD-convention dependence via gauge-equivariant local reference frames~\citep{wiersma2020cnns,haan2021gauge} and to improve inter-face geometric encoding with richer equivariant descriptors.
Another direction is replacing approximate UV-grid encodings with exact NURBS-based representations~\citep{fan2024neuronurbs} and better capturing trim geometry via boundary loops.
Exploring alternatives to standard face-adjacency graphs, such as heterogeneous face--edge--vertex graphs~\citep{li2026brephgnet}, could capture richer topological structure.
Further directions are generative B-rep modeling and extension to assemblies.

\begin{ack}
This research was supported by Flanders Make, the strategic research centre for the Flemish manufacturing industry.
\end{ack}

\bibliographystyle{apalike}
\bibliography{references}

\newpage
\appendix
\section{Computational cost and graph sizes}
\label{app:cost}
\Cref{tab:cost} compares preprocessing and runtime cost.
SO3brep reuses the ray features of FoV-Net and adds per-edge attributes, so preprocessing takes 174--243\,ms per part, against 116--195\,ms for the other B-rep models and 255--283\,ms for VN-PointNet.
It has the fewest parameters; the CG tensor products make inference and training steps $2$--$3\times$ slower than for FoV-Net and UV-Net, and about $6\times$ faster at inference, with an order of magnitude less training memory, than for VN-PointNet.

\begin{table}[h]
\centering
\caption{Computational cost.
Preprocessing is averaged over 1,000 parts per dataset.
Runtime is measured on Fusion 360 with batch size 64 on an NVIDIA A5000, mean\,$\pm$\,std over 1,000 forward passes after warm-up.}
\label{tab:cost}
\footnotesize
\begin{tabular*}{\textwidth}{@{\extracolsep{\fill}}l cc c ccc}
\toprule
& \multicolumn{2}{c}{Preprocessing (ms/part)} & & \multicolumn{3}{c}{Runtime (Fusion 360)} \\
\cmidrule(lr){2-3} \cmidrule(lr){5-7}
Model & Fusion 360 & MFTRCAD & Params (M) & Inference (ms) & Train step (ms) & Train mem.\ (MB) \\
\midrule
UV-Net & 115.7 & 195.3 & 1.36 & \ms[2.6em]{10.2}{0.1} & \ms[2.6em]{30.9}{1.4} & 854.6 \\
FoV-Net & 125.8 & 162.2 & 0.59 & \ms[2.6em]{8.3}{0.4} & \ms[2.6em]{32.0}{2.8} & 414.7 \\
VN-PointNet & 282.7 & 255.2 & 1.08 & \ms[2.6em]{128.9}{1.4} & \ms[2.6em]{419.9}{1.2} & 8701.9 \\
\midrule
\textbf{SO3brep} & 173.8 & 243.1 & 0.40 & \ms[2.6em]{22.0}{0.2} & \ms[2.6em]{85.4}{4.4} & 669.2 \\
\bottomrule
\end{tabular*}
\end{table}

\paragraph{Graph sizes and scaling.}
\Cref{tab:graphs} lists the sizes of the face-adjacency graphs.
Parts have 15--54 faces on average and at most 523.
Each message-passing layer evaluates one CG tensor product, one radial network, and one attention weight per edge, each at a fixed cost for fixed channel widths, followed by node-wise updates.
A forward pass through $L$ layers therefore costs $O(L\,(|\mathcal{E}| + |\mathcal{V}|))$ and grows linearly with the number of edges $|\mathcal{E}|$, as in any message-passing GNN.
The cost per edge depends on the channel widths and on $\ell_{\max}$ (\Cref{sec:conclusion}), not on the size of the graph.

\begin{table}[h]
\centering
\caption{Face-adjacency graph sizes.
Edges are directed: each pair of adjacent faces contributes one edge in each direction.}
\label{tab:graphs}
\footnotesize
\begin{tabular}{l r rr rr}
\toprule
& & \multicolumn{2}{c}{Faces} & \multicolumn{2}{c}{Edges} \\
\cmidrule(lr){3-4} \cmidrule(lr){5-6}
Dataset & Parts & Mean & Max & Mean & Max \\
\midrule
MFCAD++ & 59,649 & 30.0 & 86 & 306.5 & 1,096 \\
MFTRCAD & 28,637 & 25.5 & 149 & 236.3 & 1,680 \\
Fusion 360 & 35,680 & 14.7 & 421 & 130.1 & 4,428 \\
SolidLetters & 96,430 & 33.8 & 523 & 360.7 & 6,224 \\
BenDFM & 14,000 & 53.9 & 126 & 463.1 & 1,232 \\
\bottomrule
\end{tabular}
\end{table}

\end{document}